\documentclass[letterpaper, 10 pt, conference]{ieeeconf}  

\IEEEoverridecommandlockouts                              

\usepackage{graphicx}
\usepackage{amsmath}
\usepackage{algorithm}
\usepackage{algpseudocode}
\usepackage{booktabs}
\usepackage{cite}
\usepackage{hyperref}

\title{\LARGE \bf
Blind-Wayfarer: A Minimalist, Probing-Driven Framework \\ for Resilient Navigation in Perception-Degraded Environments
}

\author{Yanran Xu, Klaus-Peter Zauner, Danesh Tarapore
\thanks{Authors are with School of Electronics and Computer Science, University of Southampton, Southampton, U.K.}%
}

\begin{document}

\maketitle
\thispagestyle{empty}
\pagestyle{empty}

\begin{abstract}

Navigating autonomous robots through dense forests and rugged terrains is especially daunting when exteroceptive sensors---such as cameras and LiDAR sensors---fail under occlusions, low-light conditions, or sensor noise. We present Blind-Wayfarer, a probing-driven navigation framework inspired by maze-solving algorithms that relies primarily on a compass to robustly traverse complex, unstructured environments. In 1,000 simulated forest experiments, Blind-Wayfarer achieved a 99.7\% success rate. In real-world tests in two distinct scenarios---with rover platforms of different sizes---our approach successfully escaped forest entrapments in all 20 trials. Remarkably, our framework also enabled a robot to escape a dense woodland, traveling from 45 m inside the forest to a paved pathway at its edge. These findings highlight the potential of probing-based methods for reliable navigation in challenging perception-degraded field conditions. Videos and code are available on our website \url{https://sites.google.com/view/blind-wayfarer}

\end{abstract}

\section{INTRODUCTION}

The growing demand for ground robots in challenging, unstructured environments—such as forests \cite{mattamalaWildVisualNavigation2024,niuEmbarrassinglySimpleApproach2023, weerakoon2024vapor}, agricultural grasslands \cite{kahnBADGRAutonomousSelfSupervised2020}, and search and rescue \cite{mikiLearningWalkConfined2024}—exposes the limitations of conventional navigation systems. Most studies in autonomous navigation critically assume that their exteroceptive perception systems can reliably detect obstacles and accurately estimate the robot-obstacle relative position. However, in unstructured cluttered terrains such as in forest environments, perception-based approaches often fail \cite{kong2021avoiding, sathyamoorthy2023using,sathyamoorthyvern}. Moreover, localization methods are also affected in such environments: GPS signals degrade under dense tree canopies, and SLAM-based localization techniques are error-prone due to nondiscriminatory features with repetitive textures \cite{sahili2023survey}. The maze-like structure of many natural settings, where large fallen branches form barriers and dense undergrowth create unpredictable passageways, further complicates navigation by potentially trapping robots in complex regions.

\begin{figure}[htbp]
  \centering
    \includegraphics[width=0.95\linewidth]{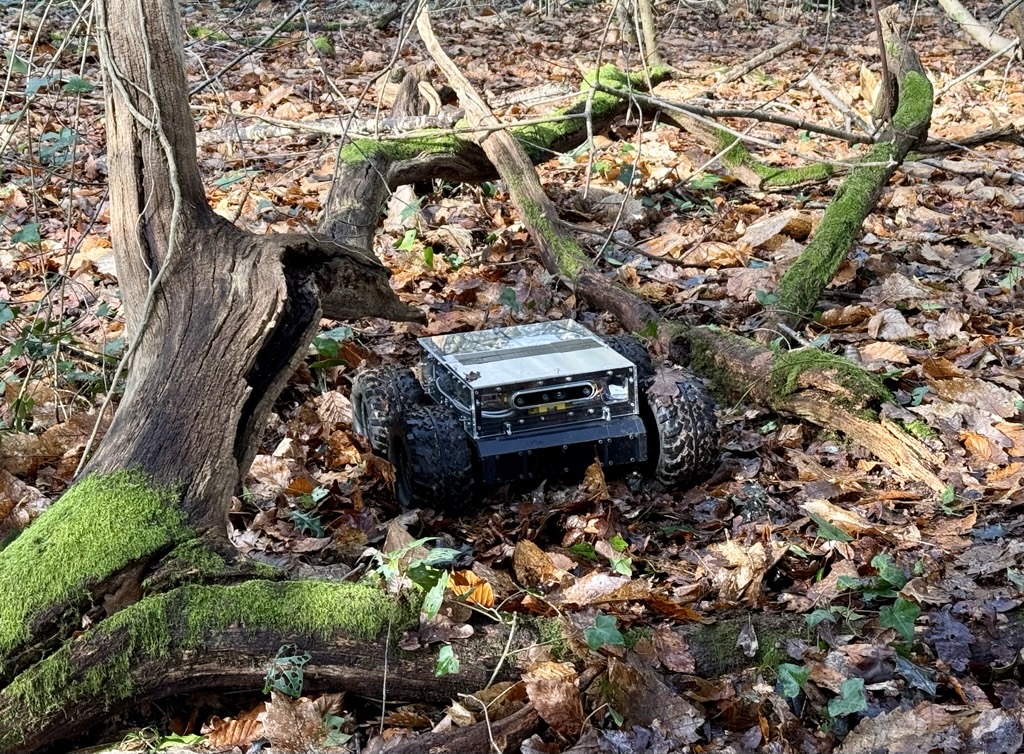}\\
  \caption{ForestRover, a portable 4WD robot in a dense obstruction trial scenario at the Southampton Common Woodlands.}
  \label{Figure:figrover}
\end{figure}

We present Blind-Wayfarer, a probing-driven navigation framework that eliminates the requirement for detailed exteroceptive perception or high-precision localization. Inspired by the classical Pledge algorithm's maze-solving principles \cite{abelson1980turtle}, our method operates with a minimal sensor suite: a compass for global heading reference, an inertial measurement Unit (IMU) for robot tilt detection, and platform-specific sensors to detect immobilization (i.e., inability of the robot to move in the desired direction). Through continual real-time feedback to physical interactions, our approach enables autonomous recovery from forest entrapment scenarios caused by unperceived obstacles. Our approach decouples navigation performance from accurate environment perception, proving particularly effective in perception-degraded environments.

The main contributions of our work are:
\begin{itemize}
    \item A probing-driven navigation framework to prevent robots from remaining trapped when obstacles cannot be reliably perceived or in perception-denied scenarios, such as night-time operations. 
    \item The design of an abstract forest simulation environment to develop and systematically evaluate our probing-driven navigation framework;
    \item Experimental validation of the proposed framework both in simulations and in real-world forest scenarios with two different rover platforms.
\end{itemize}

\section{RELATED WORK}
\label{sec:related_work}

\subsection{Perception-Based Navigation in Complex Environments}
Perception-based navigation approaches primarily utilize exteroceptive sensors, such as cameras and LiDAR sensors, and may be broadly categorized into geometry-based and appearance-based methodologies. Geometry-based approaches leverage depth information to construct obstacle maps \cite{shenAutonomousMultifloorIndoor2011} and generate collision-free trajectories, demonstrating successful navigation in structured environments \cite{shenAutonomousMultifloorIndoor2011, kimEndtoendDeepLearning2018, zhouRobustEfficientQuadrotor2019, wangLearningbased3DOccupancy2021}, vegetated outdoor terrains \cite{weerakoon2024vapor}, and limitedly in forests \cite{mattamala2024autonomous}. However, their robustness diminishes in highly unstructured and cluttered environments due to inherent limitations in perception \cite{kong2021avoiding, sathyamoorthy2023using}. Thin obstacles, such as dense vegetation and bushes, often produce incomplete or highly noisy depth readings, resulting in unreliable environmental representation. Furthermore, in confined spaces, the proximity of obstacles may fall below the minimum detection range of LiDAR or depth cameras, resulting in critical blind spots. 

Appearance-based approaches rely on visual features and texture analysis to infer traversable regions, achieving impressive performance in grasslands \cite{kahnBADGRAutonomousSelfSupervised2020, mayuku2021self} and forest environments \cite{mattamalaWildVisualNavigation2024}. However, their performance remains highly sensitive to lighting conditions. For example, rapid illumination changes or high-contrast shadows beneath forest canopies degrade feature extraction reliability, reducing the distinctiveness and repeatability of extracted features and leading to unreliable feature matching \cite{sathyamoorthyvern}. Additionally, small-sized portable ground robots operating in forest terrains frequently encounter occlusions from forest leaf litter, and dense vegetation, due to their low-viewpoint, further limiting the reliability of appearance-based approaches \cite{niuEmbarrassinglySimpleApproach2023}.

\subsection{Strategies for handling Perception Failure}
When exteroceptive sensing proves unreliable, robots are prone to repeated collisions with obstacles that remain unseen or partially occluded. Various collision-driven strategies address this issue by leveraging localization to estimate the robot’s position relative to previously encountered obstacles \cite{mulgaonkarTiercelNovelAutonomous2020, fuCouplingVisionProprioception2022, jinResilientLeggedLocal2023}.  For instance, \cite{mulgaonkarTiercelNovelAutonomous2020} maps previously collided obstacles to guide avoidance, while \cite{fuCouplingVisionProprioception2022} incorporates a collision-based fallback module to handle perception failures. Similarly, \cite{jinResilientLeggedLocal2023} employs reinforcement learning to refine avoidance policies in response to collisions. 

However, in dense, uneven forest environments, SLAM-based localization approaches are computationally expensive, and often unreliable. Multi-sensor fusion (e.g., LiDAR sensors, cameras, and IMUs) can exceed the limited resources of computationally-constrained embedded systems, hindering real-time performance. Moreover, the repetitive textures in dense vegetation offer few distinctive landmarks, challenging visual SLAM with loop closures \cite{sahili2023survey}. Additionally, fallen tree logs, branches, and moving leaf foliage frequently block exteroceptive sensors, resulting in partial scans and reduced localization accuracy, while the bumpy terrain exacerbates camera motion blur and IMU drift.


\subsection{Maze-Solving Algorithms}
A more general solution to repeated collisions or entrapment may be found in the family of classical maze-solving approaches. Early works, such as Bug1/2 \cite{lumelskyDynamicPathPlanning1986}, DistBug \cite{kamonSensoryBasedMotionPlanning1997}, and $\mu\text{Nav}$ \cite{mastrogiovanniRobustNavigationUnknown2009}, depend on the robot’s awareness of both obstacle boundaries and its relative position to the goal. In the cluttered, perception-degraded scenarios discussed above, these assumptions often fail due to the lack of reliable localization. 

The \emph{Pledge Algorithm} \cite{abelson1980turtle}, by contrast, only requires a mechanism to track accumulative turning angles, enabling a robot to systematically exit a maze without precise knowledge of its absolute position. While the original formulation emphasizes boundary-following, its minimal sensing requirements (i.e., heading changes) make it especially applicable in settings where exteroceptive data may be partial or unreliable. Consequently, revisiting and extending Pledge-inspired logic may offer a robust strategy to mitigate prolonged entrapment under real-world conditions that challenge traditional perception-driven methods.

\section{Methodology}
\label{sec:methodology}
To address challenges in perception-degraded forest environments, we propose a family of \emph{probing-driven} navigation algorithms inspired by the classic Pledge maze-solving technique. Crucially, these methods rely only on a compass sensor for heading $\psi$ and anomaly detection. In our study anomalies comprise robot flip-over risks, collisions, and robot immobilization. Below, we outline the general navigation scheme, present our \emph{Pledge-inspired} algorithms, and then highlight baseline \emph{stochastic} and \emph{reactive} approaches.

All tested algorithms share the simple control loop shown in Algorithm~\ref{general_alg}. The robot attempts to move toward a user-specified goal direction $d_{\text{goal}}$ if no anomaly is detected. Once an anomaly is detected, the robot invokes an anomaly-handling routine, which varies among algorithms.


\begin{algorithm}[t]
  \caption{Common Navigation Control Loop}
  \label{general_alg}
  \begin{algorithmic}[1]
    \State \textbf{Input:} Goal direction $d_{\text{goal}}$
    \While{Goal not reached}
      \If{\textit{no anomaly}}
        \State Move toward $d_{\text{goal}}$ \Comment{Follow goal direction}
      \Else
        \State Invoke escape handler \Comment{Alg. specific response}
      \EndIf
    \EndWhile
  \end{algorithmic}
\end{algorithm}

\subsection{Pledge-Inspired Algorithms}
\label{subsec:pledge_algs}
When large fallen branches or dense thickets form maze-like obstructions, stochastic (randomized exploration) or reactive (instantaneous sensor-driven) approaches often risk repeated collisions. By contrast, the Pledge algorithm, originally devised for maze-solving, enables agents to escape from mazes by maintaining a cumulative turning angle $C$.

\paragraph{Pledge Algorithm}
 Assuming the initial turning direction $\tau$ is clockwise (CW), the agent increments $C$ by turning angles $\theta$ for CW and decrements it for counterclockwise (CCW). When encountering a wall, the agent pivots CW; otherwise, it follows a left-hand wall-following rule until $C$ resets to zero, signaling a successful escape \cite{abelson1980turtle}.

In perception-degraded forest scenarios where ``walls'' are not reliably detected, anomalies serve as the trigger for turning CW. When no anomaly is detected and $C$ is not zero, the robot gradually probes CCW to try to follow the boundaries of the untraversable terrains; if $C$ is $0$, it follows the goal direction. The robot's turning is controlled using a simple PID controller (for details see \text{https://github.com/Xu-Yanran/blind\_wayfarer}).

The turning angle \(\theta\) is set to \(30^\circ\) in our study. Parameter sensitivity analysis reveals that our proposed approach performs significantly better with \(\theta \in [10^\circ, 50^\circ]\) than with \(\theta \in [50^\circ, 180^\circ]\), and remains robust throughout the \(10^\circ\)--\(50^\circ\) range. Consequently, \(30^\circ\) is selected as a suitable mid-range value.

\begin{algorithm}[t]
  \caption{Pledge-Inspired Escape Handler}
  \label{pledge_handler}
  \begin{algorithmic}[1]
    \State \textbf{Input:} Turn direction $\tau$ (CW or CCW), turn angle $\theta$
    \State \textbf{Initialize:} Angle counter $C \gets 0$
    \State Reverse from obstacle
    \State Turn $\tau$ by $\theta$ \Comment{Pivot on spot}
    \State $C \gets C + \theta$ \Comment{Accumulate turning angle}
    \While{$C \neq 0$} \Comment{Systematic recovery loop}
      \If{\textit{no anomaly}}
        \State Curved probing in $-\tau$ \Comment{Counter-rotation probe}
        \State $C \gets C - \Delta\psi$ \Comment{Update with change in yaw angle}
      \Else
        \State Reverse from obstacle
        \State Turn $\tau$ by $\theta$
        \State $C \gets C + \theta$
      \EndIf
    \EndWhile
  \end{algorithmic}
\end{algorithm}

\paragraph{Pledge Variants}
Table~\ref{tab:pledge_variants} summarizes \emph{ForestPledge} and its variants. \emph{Forest-Pledge} incorporates a reset mechanism for the cumulative turning angle when the rover’s heading realigns with the goal direction---an adjustment aimed at reducing superfluous recovery actions in dense forest clutter. All other proposed variants build upon Forest-Pledge but modify the \emph{initial turn direction} $\tau$ \textbf{only on the first detected anomaly} (line 6 in Alg. \ref{general_alg}) triggering the escape handler.

\noindent\textbf{Forest-Pledge}: Retaining the core Pledge procedure. If the rover’s heading already approximates the goal direction ($\pm 5^\circ$) after an anomaly, the cumulative turning angle is reset to zero.

\noindent\textbf{Rand-Pledge}: Randomizes $\tau$, selecting CW or CCW, on the first detected anomaly to eliminate bias.

\noindent\textbf{CV-Pledge}: Selects $\tau$ opposite the observed collision vector on the initial collision, steering away from the collision. In this study, the collision vector refers to whether the collision occurs on the left or right side of the robot.

\noindent\textbf{OO-Pledge}: If the obstacle’s orientation is approximately aligned with the rover’s goal heading ($\pm 20^\circ$), $\tau$ follows that orientation; otherwise, it defaults to Forest-Pledge behavior.

\noindent\textbf{Levy-Pledge}: Mitigates dense probing by adding stochasticity to the turn magnitude by sampling $\theta \sim \text{Lévy}(0,1)$, clipped to range $[10^\circ, 50^\circ]$.

\begin{table}[t]
  \centering
  \caption{Parameters for Pledge-Inspired Algorithms}
  \label{tab:pledge_variants}
  \begin{tabular}{@{}l c c@{}}
  \toprule
  \textbf{Algorithm} & 
  \begin{tabular}[c]{@{}c@{}}
    Initial 
    Turning Direction ($\tau$)
  \end{tabular} &        
  \begin{tabular}[c]{@{}c@{}}
    Turning Angle 
    ($\theta$)
  \end{tabular} \\
  \midrule
  Pledge & Fixed (CW) & 30$^\circ$ \\
  Forest-Pledge & Fixed (CW) & 30$^\circ$ \\
  Rand-Pledge & $\tau \sim \text{Uniform}\{\text{CW}, \text{CCW}\}$ & 30$^\circ$ \\
  CV-Pledge & Opposite collision vector & 30$^\circ$ \\
  OO-Pledge & Aligned to obstacle orientation & 30$^\circ$ \\
  Levy-Pledge & Fixed (CW) & 
  \begin{tabular}[c]{@{}c@{}}
    $\theta \sim \text{Lévy}(0,1)$ 
  \end{tabular} \\
  \bottomrule
  \end{tabular}
\end{table}


\subsection{Baseline Algorithms}

To establish a rigorous benchmark for evaluating our proposed navigation framework, we implement four baseline strategies representative of \textbf{stochastic} and \textbf{reactive} paradigms. These methods are chosen to reflect common solutions for perception-degraded environments where precise perception or accurate localization is infeasible.

\noindent\textbf{Stochastic Algorithms}

\noindent\textit{Levy Walk}: The rover proceeds straight toward $d_{\text{goal}}$ until abnormal events. It then samples a new heading $\theta$ from a uniform distribution and a step size from a L\'{e}vy distribution, enabling sufficient exploration of the environment.

\noindent\textit{CV-Levy}: A hybrid variant that considers both a L\'{e}vy-distributed step size and a \emph{collision-vector} to bias the robot's heading away from the obstacle.

\noindent\textbf{Reactive Algorithms}

\noindent\textit{Collision Vector React (CV-React)}: If an anomaly is detected, the rover steers away from the estimated collision vector to find a likely free direction.

\noindent\textit{Obstacle Orientation React (OO-React)}: If the obstacle’s orientation appears compatible with $d_{\text{goal}}$, the robot follows it; otherwise, it turns in the opposite direction of the collision. 


\subsection{Robot Platform}
Two rover platforms, \emph{ForestRover} and \emph{MiniRover}, are employed for assessing the performance of our algorithms in cluttered, uneven forest terrain (see Table~\ref{tab:robot_specifications}). ForestRover, based on the Lynxmotion A4WD3 chassis, uses a CMPS14 IMU for heading and pitch measurements to avoid flip-over, while a Pimoroni PMW3901 optical flow sensor monitors ground-relative motion to detect immobilization. Although equipped with a RealSense camera, it remains unused to maintain a minimalist sensing approach. In contrast, the smaller, lightweight MiniRover (MonsterBorg chassis) relies on the same IMU for flip-over avoidance and uses a Raspberry Pi camera to detect immobilization by comparing successive images.

\begin{table}[t]
    \centering
    \caption{Robot Platform Specifications}
    \label{tab:robot_specifications}
    \begin{tabular}{lcc}
        \toprule
        \textbf{Robot Platform} & \textbf{ForestRover} & \textbf{MiniRover} \\
        \midrule
        \textbf{Base Chassis} &  
        \begin{tabular}[c]{@{}c@{}}
            Lynxmotion \\
            A4WD3 \cite{lynxmotion_a4wd3}
        \end{tabular}        
        & MonsterBorg \cite{piborg_monsterborg} \\
        \textbf{Weights} & 6.1\,kg & 1.9\,kg \\
        \textbf{Dimensions} & 437\,$\times$\,377\,$\times$\,228\,mm & 180\,$\times$\,160\,$\times$\,80\,mm \\
        \textbf{Ground Clearance} & 41\,mm & 30\,mm \\
        \textbf{Onboard Computer} & Raspberry Pi 5 & Raspberry Pi 3b \\
        \textbf{Motors} & 4 $\times$ DC (11.5\,kg$\cdot$cm) & 4 $\times$ DC (1.02\,kg$\cdot$cm) \\
        \textbf{Max. Velocity} & 1.4\,m/s (theoretical) & 1.0\,m/s (theoretical) \\
        \textbf{IMU} & CMPS14 & CMPS14 \\
        \textbf{Optical Flow Sensor} & Pimoroni PMW3901 & -- \\
        \textbf{Camera} & 
            \begin{tabular}[c]{@{}c@{}}
            RealSense d455i \\
            (unused)
            \end{tabular} 
        &   \begin{tabular}[c]{@{}c@{}}
            Pi Camera \\
            Module V2
            \end{tabular} \\
        \bottomrule
    \end{tabular}
\end{table}

\begin{figure}[htbp]
  \centering
  \includegraphics[width=0.48\linewidth]{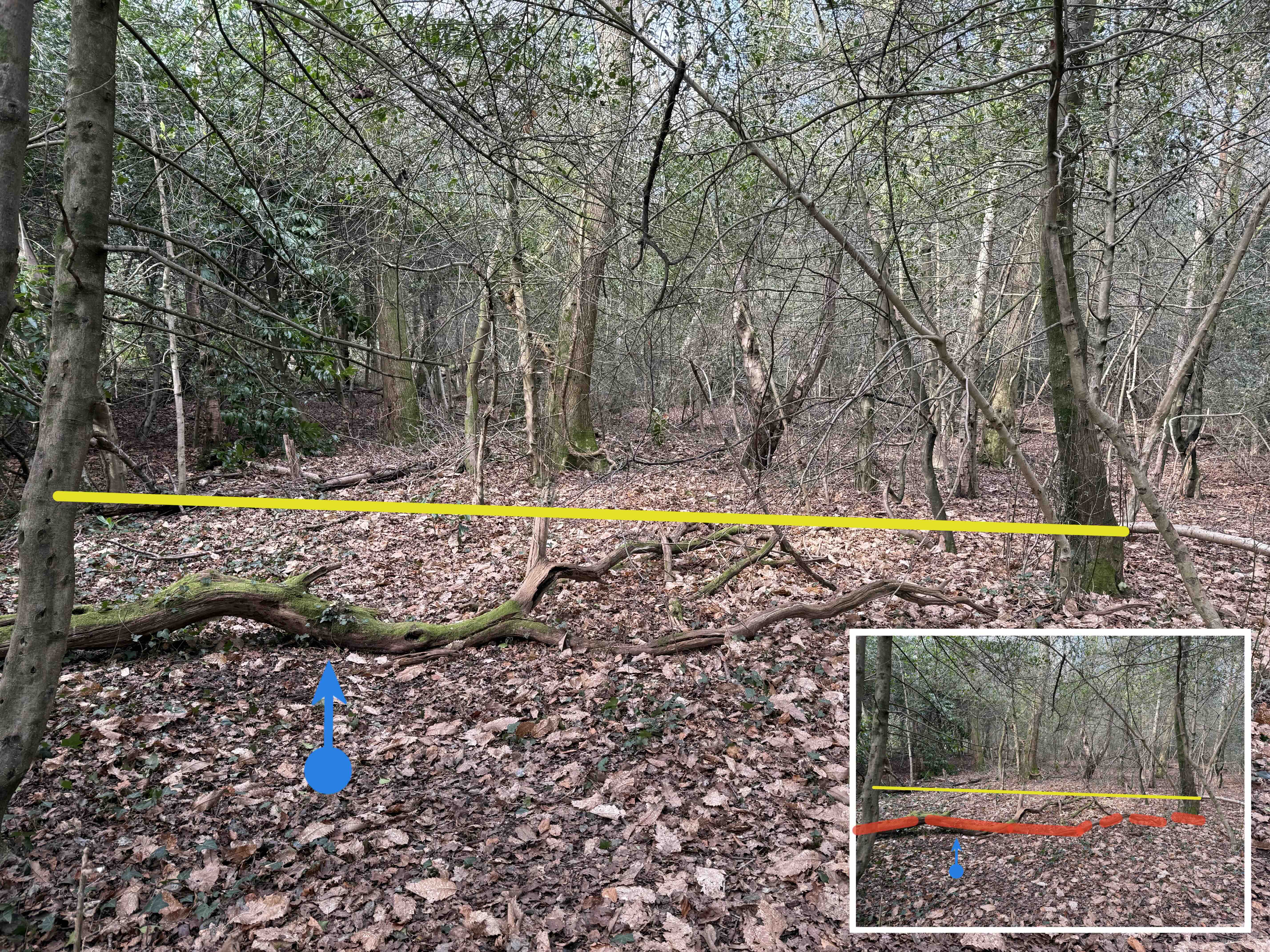}
  \includegraphics[width=0.48\linewidth]{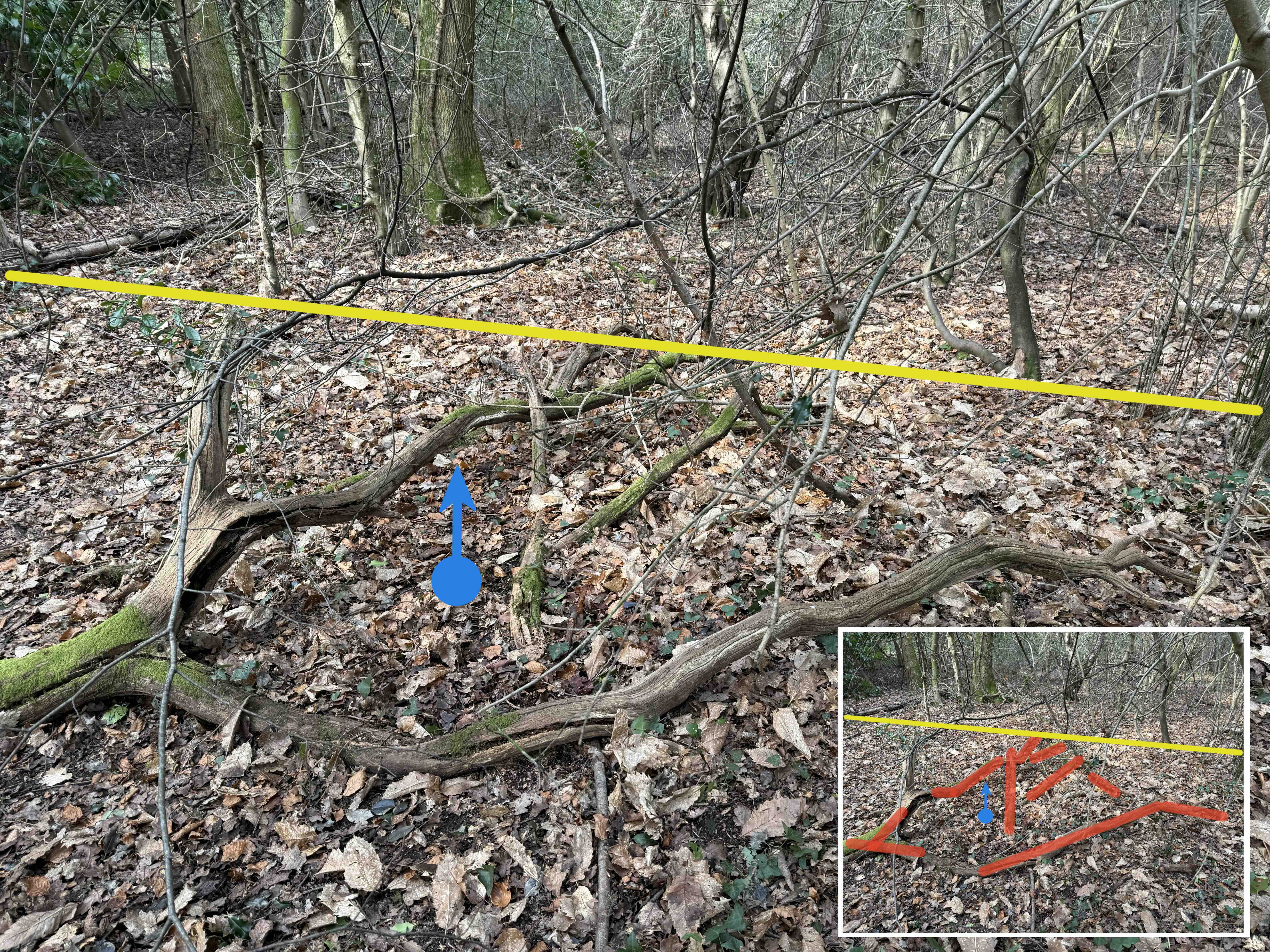}
  \caption{Two trial scenarios in Southampton Common: Extended Maze Trial, featuring large fallen branches that form barrier-like obstructions (Left); Dense Obstruction Trial, characterized by the rover being surrounded by tightly cluttered fallen branches (Right). In both scenarios, obstacles do not fully block potential paths; ForestRover can navigate some of these gaps, while the more compact MiniRover can traverse even more.  The blue dot and arrow marks the starting point and initial orientation of the robot, while the yellow line indicates the destination line. The inset image highlights the main obstacles with red lines. Both scenarios evaluate the rover’s ability to navigate cluttered, perception-degraded forest environments.}
  \label{Figure:trial_scenario_evaluation}
\end{figure}

\section{EXPERIMENTS}

\subsection{Abstract Forest Simulations}
\label{sec:simulation_setup}
\begin{figure}[t] 
    \centering
    \begin{minipage}{0.5\columnwidth} 
        \centering
        \includegraphics[width=\linewidth]{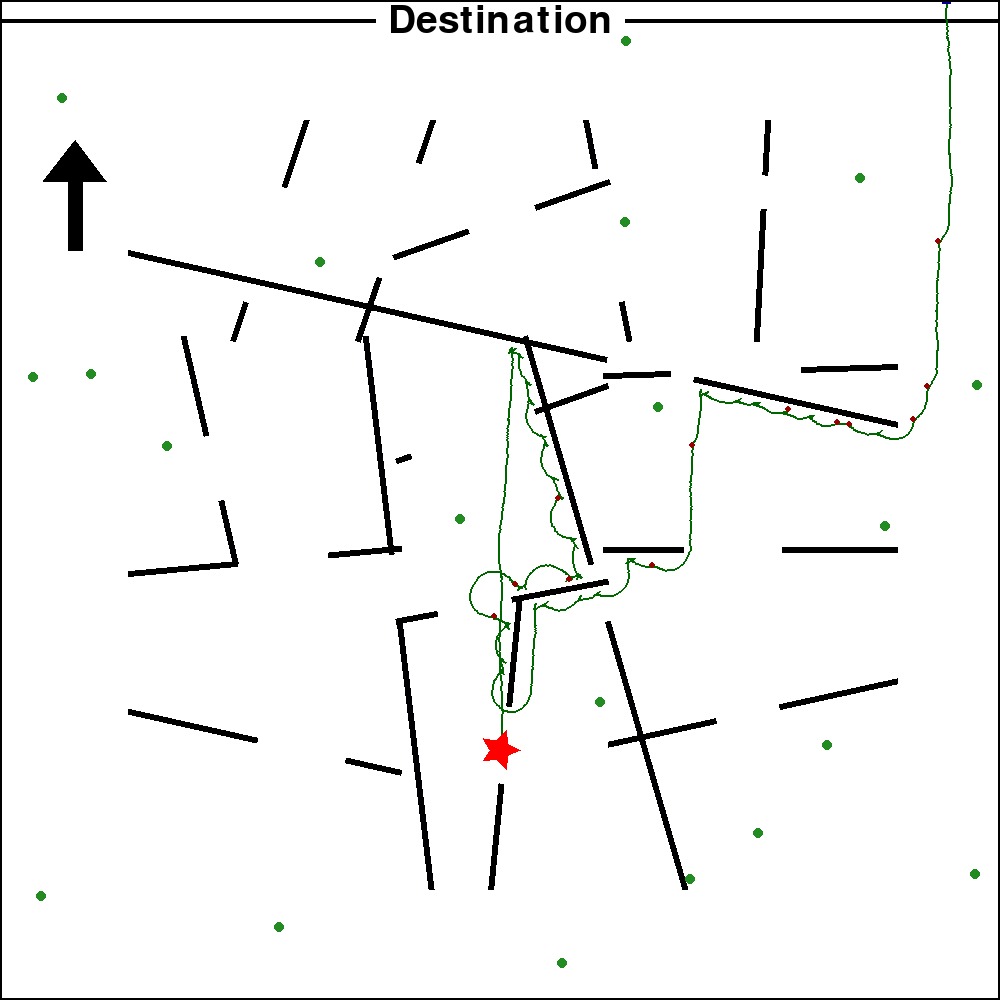}
    \end{minipage}
    \begin{minipage}{0.3\columnwidth} 
        \centering
            \includegraphics[width=0.4\columnwidth]{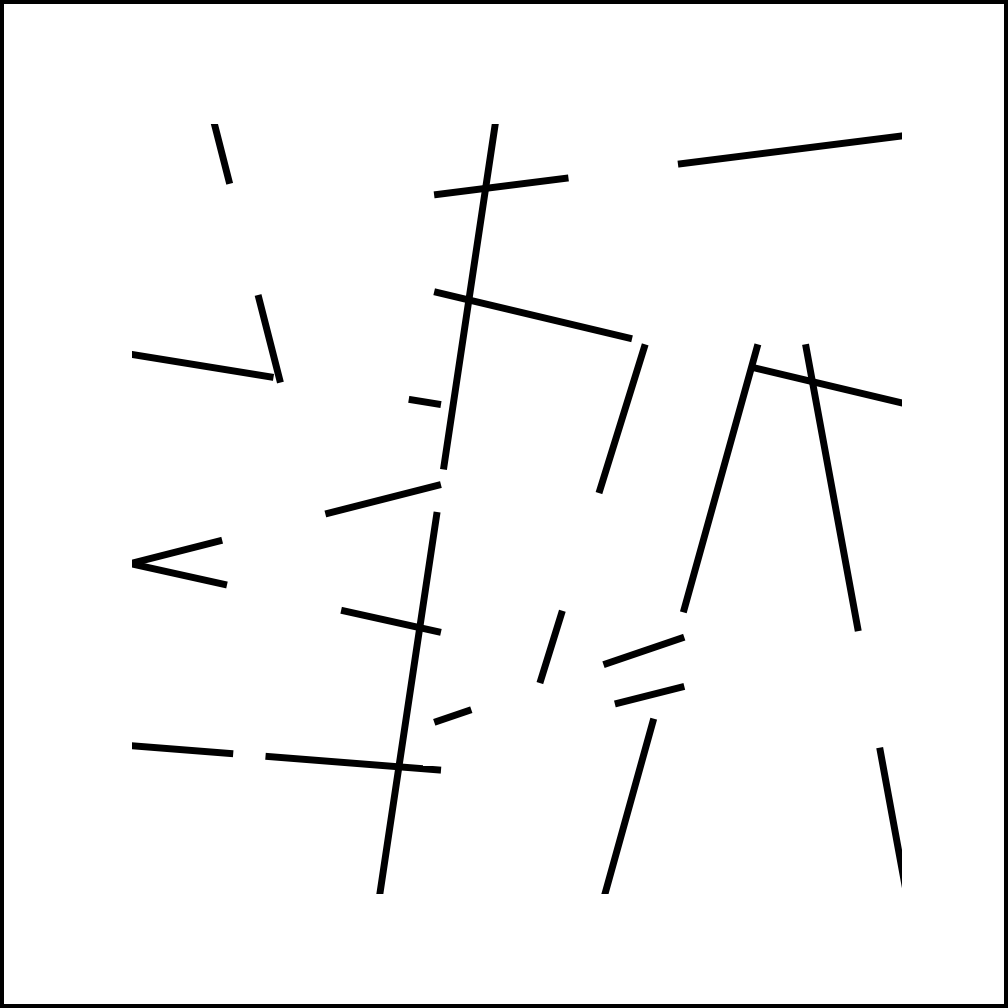}
            \includegraphics[width=0.4\columnwidth]{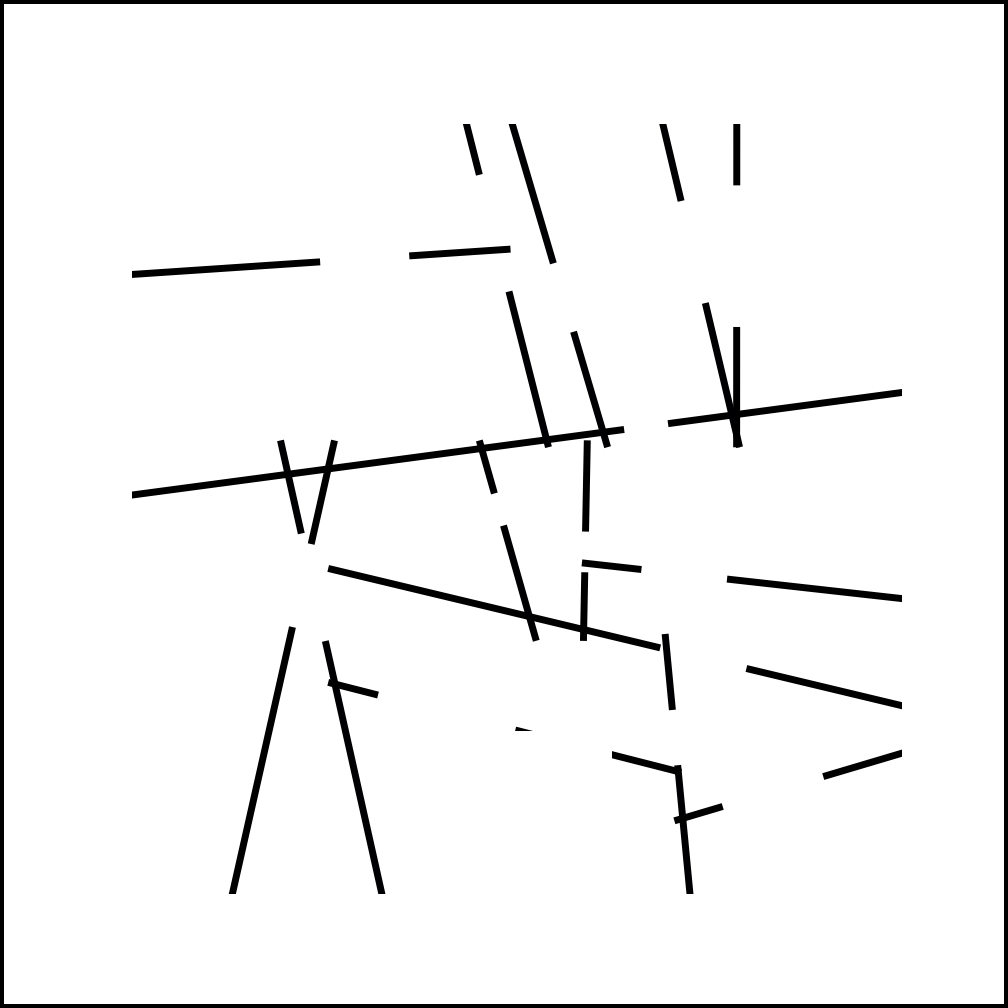} \\[1.5pt]
            \includegraphics[width=0.4\columnwidth]{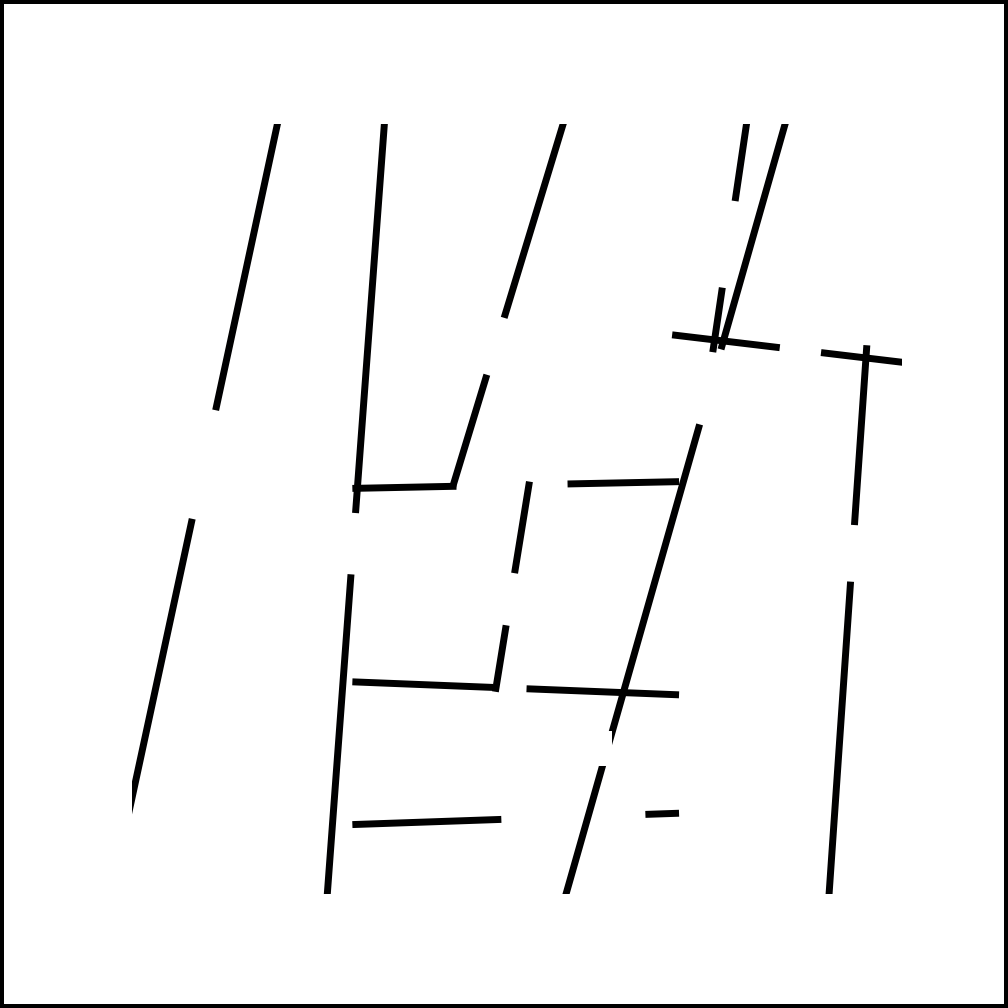}
            \includegraphics[width=0.4\columnwidth]{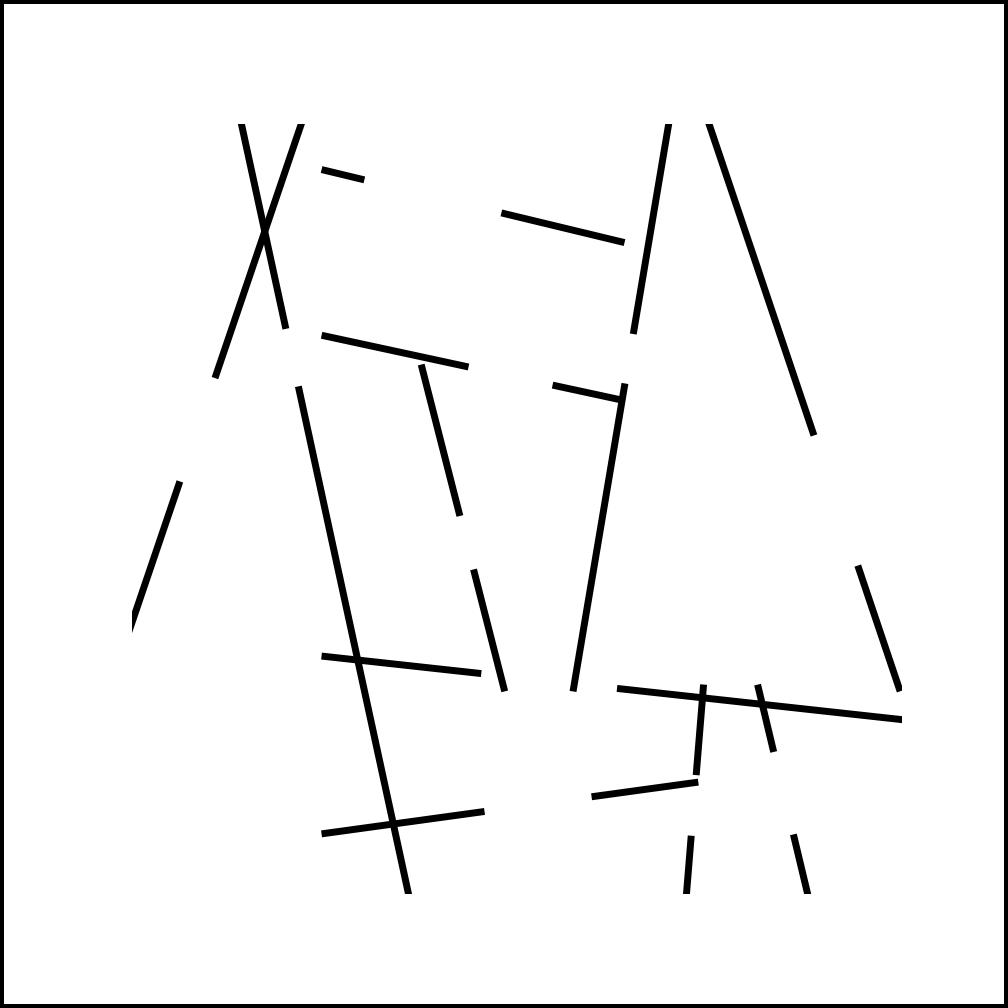} \\[1.5pt]
            \includegraphics[width=0.4\columnwidth]{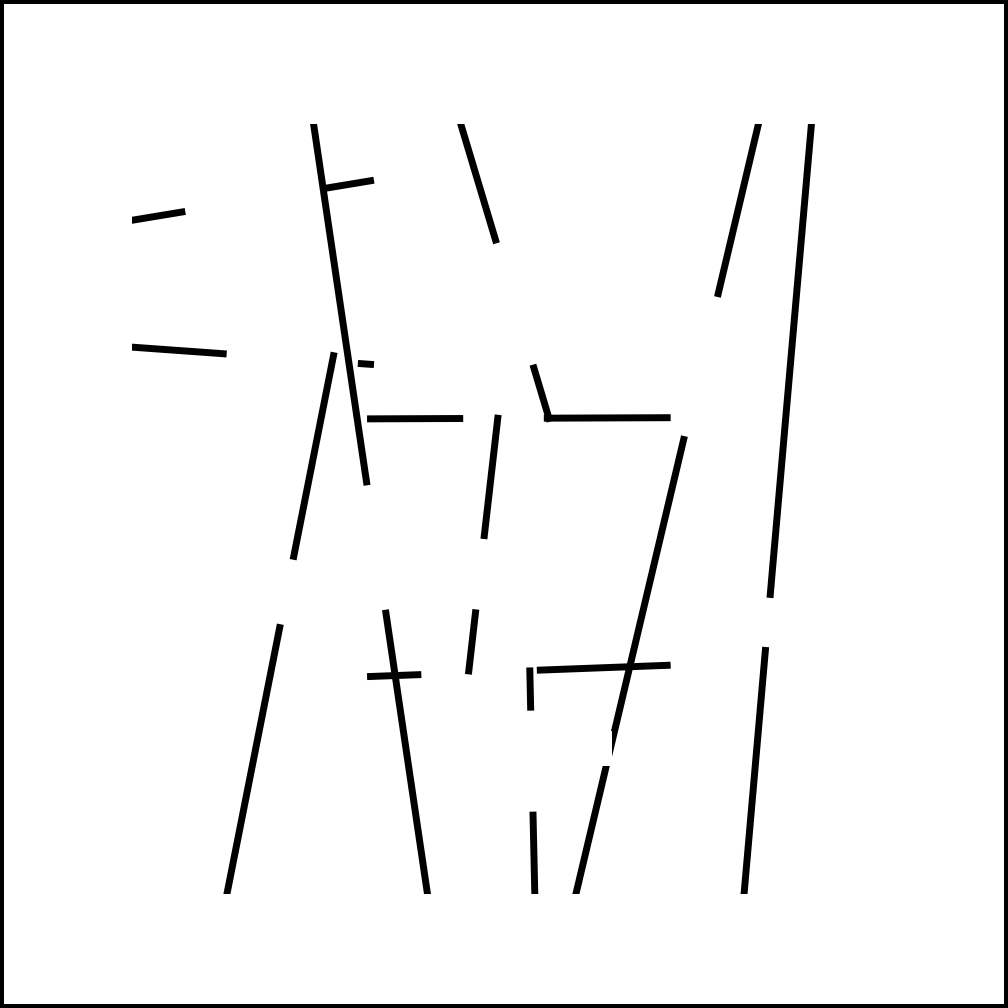}
            \includegraphics[width=0.4\columnwidth]{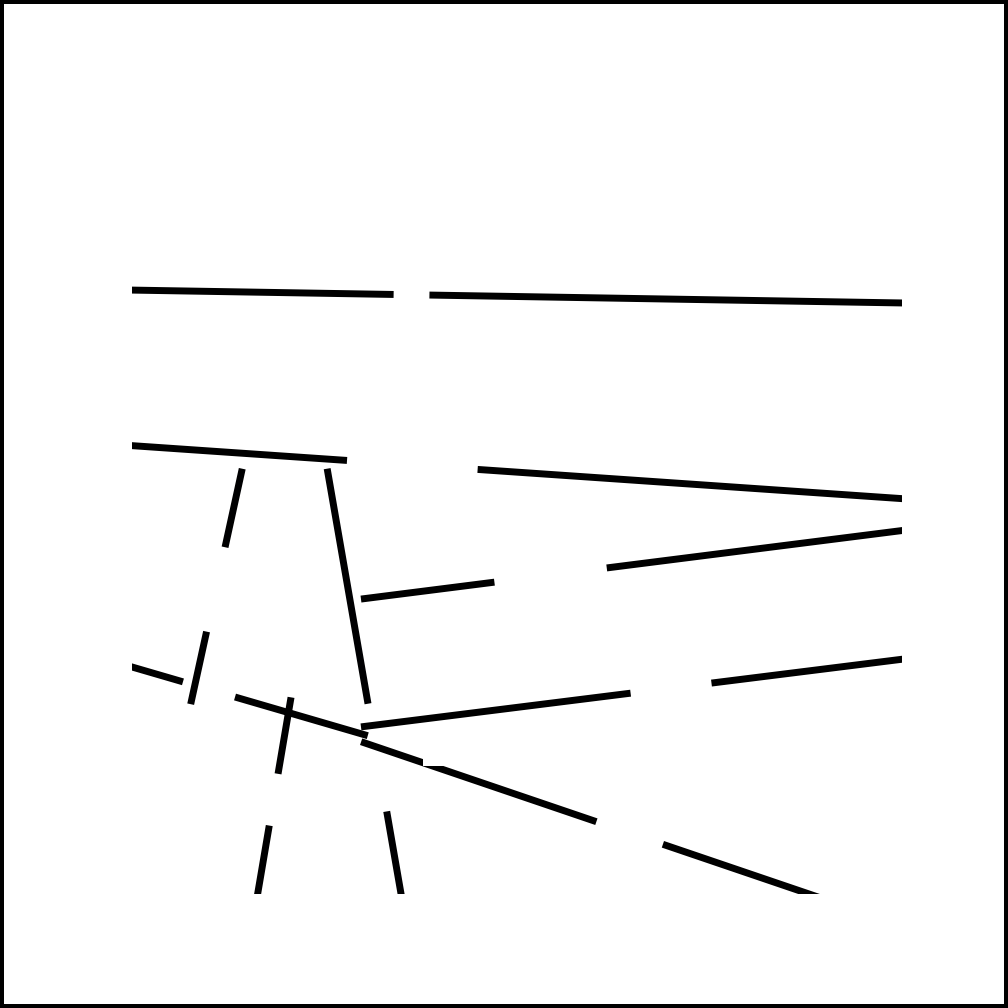} \\[1.5pt]
            \includegraphics[width=0.4\columnwidth]{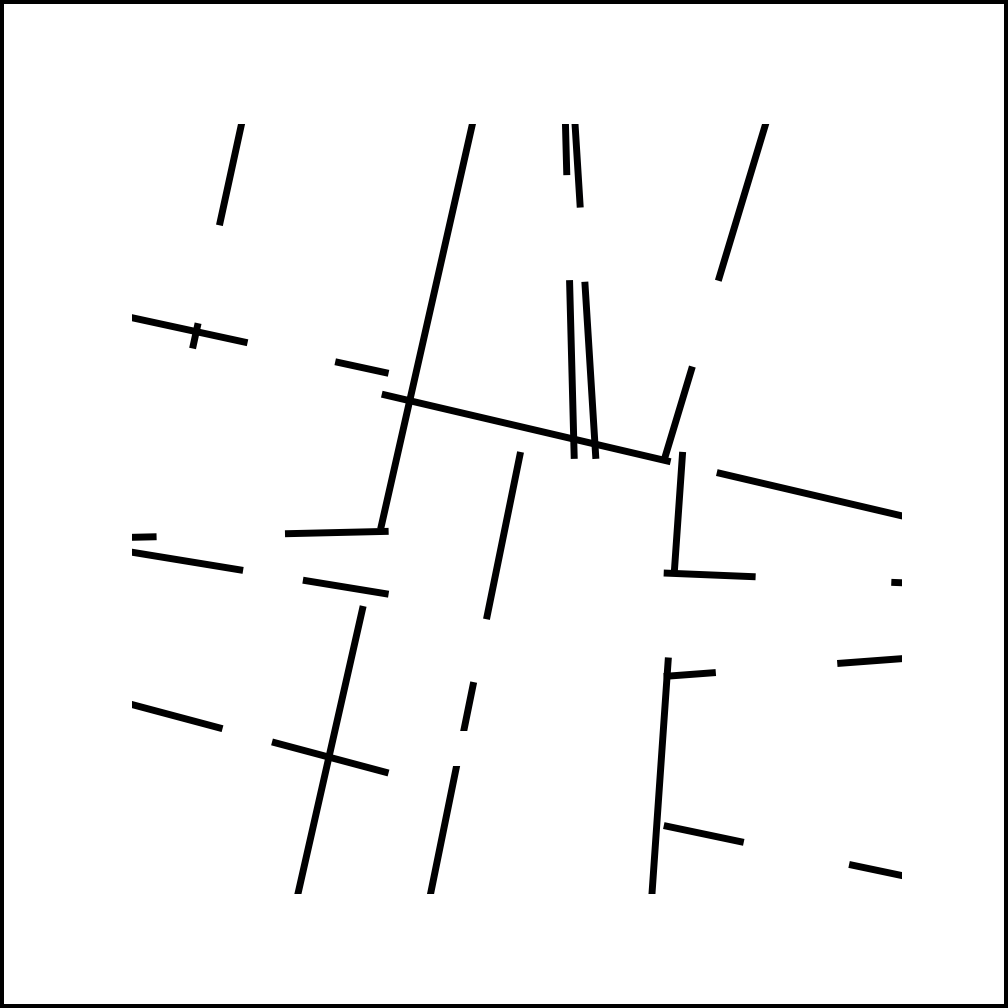}
            \includegraphics[width=0.4\columnwidth]{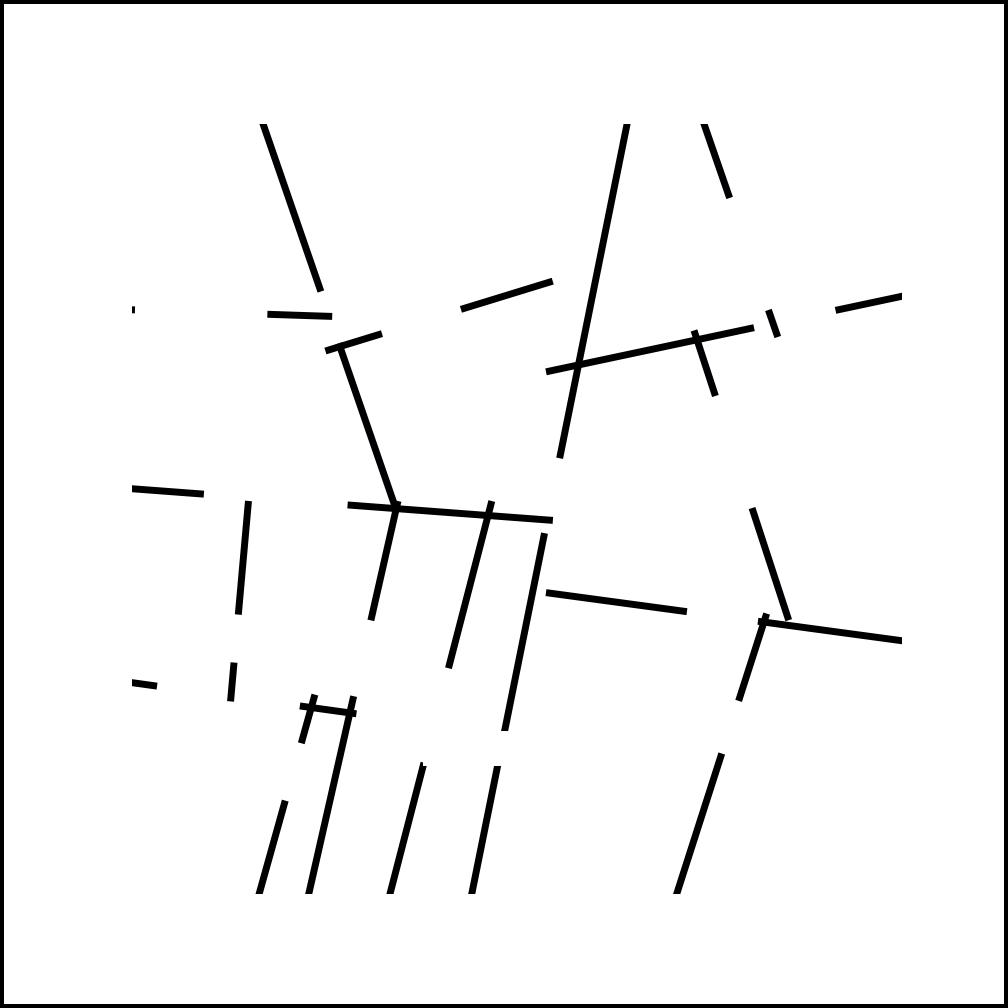}
    \end{minipage}
    \caption[Simulation environments]{
    Representative examples from the 500 abstract forest simulation environments used in our study. 
        Each environment spans a \(50\times50\,\text{m}^2\) area with a tree density of \(80\,\text{trees/hectare}\), 
        where each tree has a radius of \(0.25\,\text{m}\). 
        In the larger panel (Left), black lines indicate fallen branches, green circles represent standing trees, 
        and red dots mark locations where the robot becomes immobilized or slips on rough terrains accidentally. 
        The green line shows the robot's trajectory. The red star indicates the starting point 
        and the black arrow denotes the designated heading. 
        The smaller panels (right) illustrate additional scenarios that capture diverse obstacle layouts.
        }
    \label{Figure:sim_env}
\end{figure}

To systematically evaluate the proposed navigation framework, we develop a two-dimensional abstract environment (Fig. \ref{Figure:sim_env}) that encapsulates key challenges of forest terrains. Specifically, the environment models comprise large fallen branches---elongated obstacles that can trap or block the robot, standing trees---cylindrical obstacles that the robot must navigate around, muddy forest terrain---regions that result in wheel slippage or immobilization, where distance traveled between such events is modeled with a negative exponential distribution as $d \sim \text{Exp}(0.5)$ m.

The robot is modeled as a $0.4 \times 0.4\,\text{m}^2$ rectangle, closely matching the footprint of our ForestRover (Table \ref{tab:robot_specifications}) to minimize the simulation-to-reality gap---although parameter sensitivity analysis indicates that performance remains robust for robot footprints ranging from $0.2 \times 0.2\,\text{m}^2$ to $0.8 \times 0.8\,\text{m}^2$. In our simulation experiments, noise is added to the robot steering control and compass sensor measurements. 

\begin{figure*}[htbp]
  \centering
  \includegraphics[width=1.0\linewidth]{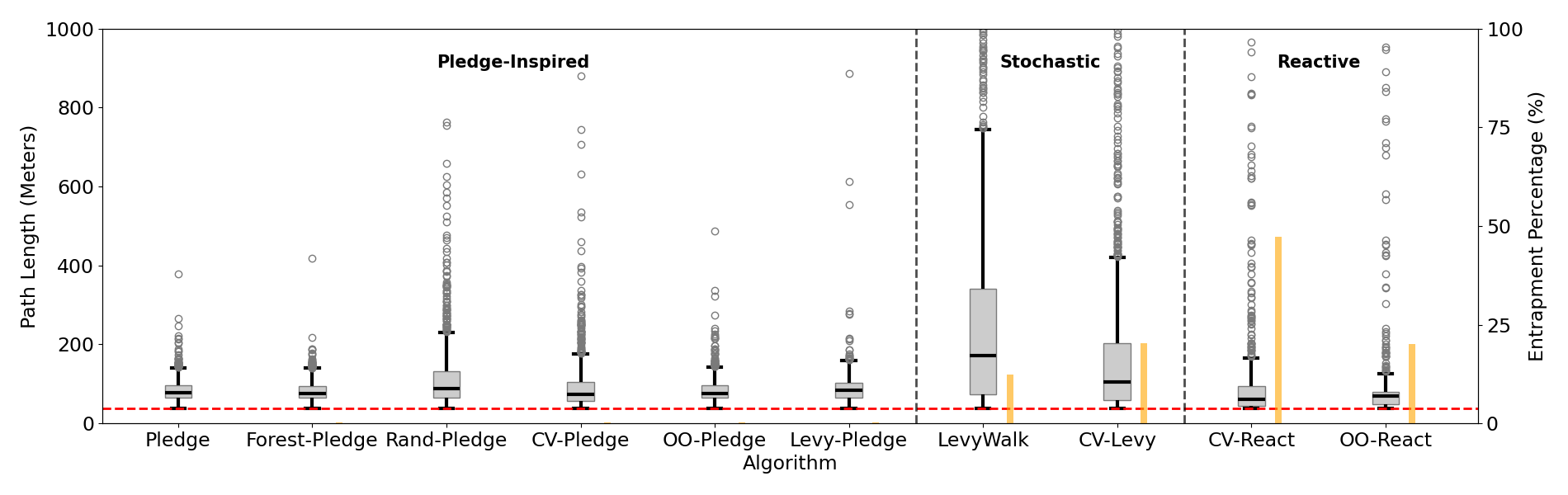}
  \caption[Simulation results]{
    Box plots of path length (left vertical axis) for each algorithm, with outliers shown as circles. 
    The orange bars represent the proportion of simulations in which the robot fails to escape from the maze-like scenario (right vertical axis), 
    and the red dashed line marks the shortest distance to the goal. 
    The algorithms are grouped into \emph{Pledge-Inspired}, \emph{Stochastic}, and \emph{Reactive} categories 
    (separated by vertical dashed lines). 
    The proposed Pledge-Inspired approaches achieve generally shorter path lengths and lower entrapment proportions, 
    indicating more reliable navigation performance in perception-degraded forest simulations 
    compared to purely stochastic or reactive baselines.
  }
  \label{Figure:sim_evaluation}
\end{figure*}


We evaluate stochastic, reactive, and pledge-inspired navigation algorithms across 500 randomly generated forest environments (Fig.~\ref{Figure:sim_env}), assessing their ability to reach a fixed destination line without permanent entrapment. A destination line is used rather than a single point to evaluate performance as precise localization is infeasible under perception-degraded conditions. Two performance metrics are employed to measure reliability and efficiency:

\textbf{Entrapment cases:} A trial is classified as failed if the robot is unable to to reach the destination line within 90\,s.

\textbf{Path length to destination line:} For successful trials, we measure the total distance traveled until the destination line is crossed. Trials in which the robot becomes trapped are excluded from the path-length statistics. If a trial ends with the robot exiting the environment bounds, it is considered a successful escape, and the recorded path length includes the remaining distance to the destination line.

Fig.~\ref{Figure:sim_evaluation} presents a comparative analysis of three algorithmic classes: \emph{pledge-inspired}, \emph{stochastic}, and \emph{reactive} approaches. The results indicate that \textbf{pledge-inspired} methods significantly reduce the likelihood of indefinite entrapment compared to stochastic and reactive baselines. This suggests that systematically accounting for accumulative turning angles, as done in Pledge-inspired algorithms, enhances escape robustness in complex environments.

Among pledge-inspired methods, \textit{Pledge}, \textit{Forest-Pledge}, and \textit{Levy-Pledge} demonstrate comparable performance in terms of both escape success rate and path efficiency, with path lengths of 82.5 ${\pm}$ 30.8, 81.3 ${\pm}$ 29.3, and 88.3 ${\pm}$ 46.3 meters (mean ${\pm}$ SD), respectively. By contrast, \textbf{stochastic} and \textbf{reactive} approaches exhibit greater variability in performance. The LevyWalk, for instance, while occasionally yielding shorter escape paths, suffers from high variance (247.5 ${\pm}$ 223.7 meters), with numerous trials resulting in extreme path lengths. Additionally, these methods exhibit a higher proportion of failed escapes (yellow bars in Fig.~\ref{Figure:sim_evaluation}), reinforcing the advantage of structured escape strategies over reactive or stochastic approaches.


\begin{figure}[htbp]
  \centering
  \includegraphics[width=\linewidth]{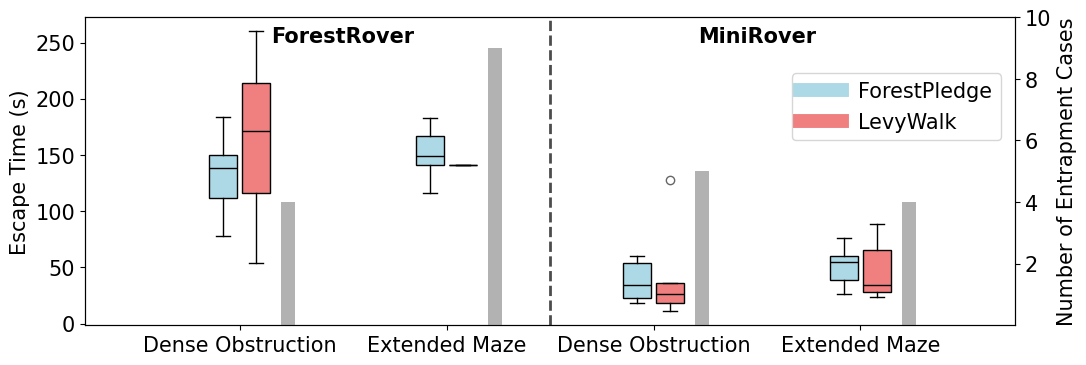}
  \caption{
      Box plots of escape time (left vertical axis) for the ForestRover (left) and MiniRover (right) 
    in two challenging forest scenarios: Dense Obstruction Trial and Extended Maze Trial. 
    The blue boxes correspond to ForestPledge and the red boxes to LevyWalk. 
    Gray bars (right vertical axis) show the number of trials in which the robot became entrapped (ForestPledge experienced no entrapment).
    Results indicate that ForestPledge generally yields lower escape times 
    and fewer entrapment occurrences than LevyWalk, 
    underscoring its robustness in complex, perception-degraded environments.
  }
  \label{Figure:compare_all}
\end{figure}

\subsection{Real Forest Experiments}
We evaluated the proposed \emph{Forest-Pledge} algorithm against a baseline \emph{LevyWalk} strategy on two platforms---the larger, more robust \emph{ForestRover} and the smaller, low-cost, light-weight \emph{MiniRover}---under two challenging trial scenarios (Fig. \ref{Figure:trial_scenario_evaluation}). \emph{Extended Maze} features large fallen branches that form barrier-like obstructions while \emph{Dense Obstruction} is characterized by the rover being surrounded by tightly cluttered fallen branches. Fig.~\ref{Figure:compare_all} summarizes the performance across the four combinations of robot and scenario. Notably, due to inertia, rovers sometimes climbed onto fallen branches during experiment trials, leading to flip-overs or loss of all-wheel contact with the ground. Human interventions averaging 11 across all 10 trials for the ForestRover (6 for MiniRover) were therefore necessary to restore mobility (for details, see supplementary figures on our website). Following the intervention, the rovers were placed near their stuck location with the same orientation, ensuring consistent experimental conditions. 

\textbf{Trial Scenarios}
In the \emph{Extended Maze} trials, \emph{Forest-Pledge} achieved a 100\% success rate, escaping in \(152 \pm 21\,\text{s}\) on ForestRover and \(51 \pm 15\,\text{s}\) on MiniRover. By contrast, \emph{LevyWalk} escaped only once out of the ten trials on ForestRover (\(141 \pm 0\,\text{s}\)) and six times on MiniRover (\(47 \pm 25\,\text{s}\)). In the \emph{Dense Obstruction} trial, \emph{Forest-Pledge} completed all runs, taking \(133 \pm 32\,\text{s}\) on ForestRover and \(38 \pm 16\,\text{s}\) on MiniRover, while \emph{LevyWalk} succeeded in six out of ten runs for ForestRover (\(164 \pm 70\,\text{s}\)) and five out of ten for MiniRover (\(46 \pm 43\,\text{s}\)). These results demonstrate that \emph{Forest-Pledge} consistently prevents entrapment, while \emph{LevyWalk} often fails to escape the presented trial scenarios. Interestingly, the larger escape times observed on ForestRover reflect its inability to fit through narrow gaps that MiniRover can traverse. MiniRover’s smaller footprint allows it to bypass obstacles more easily, thereby reducing its overall escape time. Consequently, MiniRover typically exits scenarios faster than ForestRover, especially in the \emph{Dense Obstruction} trial featuring tight passages.


\begin{figure}[htbp]
  \centering
  \includegraphics[width=1.0\linewidth]{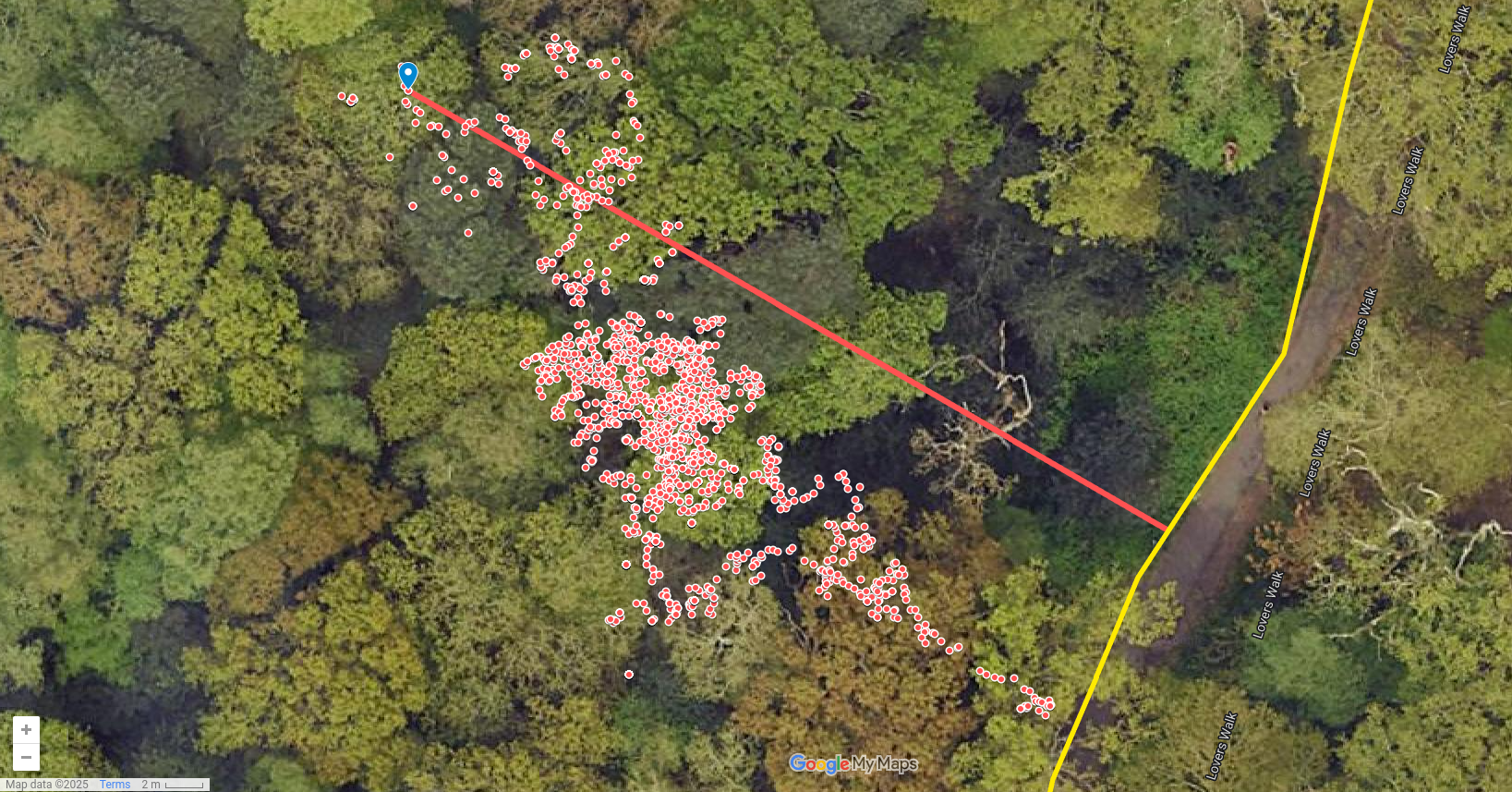}
  \caption{
    Forest Escape Challenge is illustrated on Google Satellite Maps \cite{googlemaps}. The rover begins at the top-left corner (50.93598303$^\circ$ N, 1.400852342$^\circ$ W), with its initial heading oriented perpendicular to the destination line—a paved pathway highlighted in yellow. The shortest distance between the start point and the destination line is 45 meters, represented by a solid red line. The rover’s GPS-recorded trajectory, depicted as red dots, traces its movement during the trial.
  }
  \label{Figure:initial_area}
\end{figure}

\textbf{Forest Escape Challenge}
To further validate our method in forest environments over extended distances, the rover was tasked with traveling from a designated start point in the forest to a paved pathway along the outskirts, with the shortest distance to the pathway being approximately 45 meters. Fig.~\ref{Figure:initial_area} depicts the experiment area, highlighting both the start position and the destination line, and shows the trajectory logged via GPS. Despite encountering natural obstacles, the rover completed its journey without becoming indefinitely trapped. This outcome illustrates how the \emph{Forest‐Pledge} strategy generalizes to real‐world forested environments, offering robust recovery capabilities in the face of complex terrain.

\section{DISCUSSION}
\label{sec:discussion}
We present \textbf{Blind-Wayfarer}, a minimalist navigation framework enabling robust traversal in perception-degraded environments through probing-driven interaction and systematic entrapment recovery. By leveraging the principles of the Pledge algorithm, the framework eliminates dependency on exteroceptive perception, relying primarily on a compass sensor for heading correction. The algorithm employs an inherently explainable mechanism, where navigation decisions and recovery actions are directly linked to observable interactions with obstacles and systematic heading correction rules. It is robust to rover terrain interaction-induced noise, ensuring reliable operation even in highly uneven and cluttered environments. Evaluated across 1,000 abstract forest simulations, our proposed Forest-Pledge algorithm achieved a 99.7\% success rate, outperforming stochastic and reactive baselines in both escape reliability and efficiency. Real-world validation on two rover platforms of different sizes demonstrated consistent success across 20 trials in our selected forest scenarios, and the forest escape challenge.

The Blind-Wayfarer offers distinct advantages for robotic applications, featuring its low computational costs and minimal sensor dependencies, making it suitable for swarm robot deployments \cite{tarapore2020sparse, wu2024proof, wu2024secure}. Furthermore, it provides a robust fail-safe mechanism for advanced navigation systems \cite{kahnBADGRAutonomousSelfSupervised2020, mattamalaWildVisualNavigation2024, niuEmbarrassinglySimpleApproach2023}, preventing indefinite entrapment in scenarios where primary sensors fail due to accidental hardware damage, occlusions, or low-light conditions that degrade perception.

In future work, we aim to enhance the framework's efficiency by incorporating adaptive turning strategies informed by local interaction cues. Specifically, the turning angle can be reduced in the likelihood of narrow gaps to exploit immediately available paths, while broader turns can be triggered when no likely viable paths are detected in the immediate vicinity, encouraging more exploratory behavior to break free from extended entrapment. Such a dynamic adjustment would enable a better balance between local exploitation of clear paths nearby and global exploration to escape complex obstacles. Overall, Blind-Wayfarer provides a reliable, explainable, and sensor-efficient navigation approach, demonstrating strong potential for robust traversal in perception-degraded environments.



\addtolength{\textheight}{-12cm}   









\bibliographystyle{IEEEtran}
\bibliography{main}

\end{document}